\documentclass{article}
\usepackage{spconf,amsmath,graphicx}
\usepackage{multirow,float,makecell}

\title{Deep Feed-forward Sequential Memory Networks for Speech Synthesis}
\name{Mengxiao Bi, Heng Lu, Shiliang Zhang, Ming Lei, Zhijie Yan}
\address{Alibaba Inc., Beijing, China\\
\texttt{\{mengxiao.bmx, h.lu, sly.zsl, lm86501, zhijie.yzj\}@alibaba-inc.com}}
\begin{document}
%
\maketitle
\begin{abstract}
The Bidirectional LSTM (BLSTM) RNN based speech synthesis system is among the best parametric Text-to-Speech (TTS) systems in terms of the naturalness of generated speech, especially the naturalness in prosody. However, the model complexity and inference cost of BLSTM prevents its usage in many runtime applications. Meanwhile, Deep Feed-forward Sequential Memory Networks (DFSMN) has shown its consistent out-performance over BLSTM in both word error rate (WER) and the runtime computation cost in speech recognition tasks. Since speech synthesis also requires to model long-term dependencies compared to speech recognition, in this paper, we investigate the Deep-FSMN (DFSMN) in speech synthesis. Both objective and subjective experiments show that, compared with BLSTM TTS method, the DFSMN system can generate synthesized speech with comparable speech quality while drastically reduce model complexity and speech generation time.

\end{abstract}
\section{Introduction}
In recent years, Deep Neural Networks (DNNs) have became dominant in the parametric speech synthesis back-end modeling\cite{hmm,dnn}. Compared with conventional HMM-GMM methods training at state level, neural network based methods model and predict at much smaller step, e.g. frame or even sample level\cite{dnn,blstm,wavenet}. Recurrent Neural Networks (RNNs) have shown their advantage in modeling long-term dependencies in sequential data due to their recurrent connections over time \cite{rnnlm}. And amongst all variants of recurrent neural networks, Long Short-Term Memory (LSTM) and Bidirectional LSTM (BLSTM)\cite{lstm}, alleviate the gradient vanishing problem with back-propagation through time (BPTT)\cite{bptt,vanish} training in basic RNNs, which are consequently better to learn very long-range dependencies between sequential samples. With more attention paid, LSTM/BLSTM-RNN based acoustic sequence modeling has turned out to be a great success in both speech recognition\cite{lstm-asr-1,lstm-asr-2} and speech synthesis\cite{blstm}.

Different to the basic fully-connected feed-forward neural network (FNN) that maps a fixed input within a small context window to a fixed output, DFSMN is able to capture information in a very long context by using memory blocks with look-back and look-ahead filters in a hierarchical structure. More importantly, DFSMN preserves the feed-forward structure so that it can be learned in a more efficient and stable way than BLSTM-RNN. In this paper, we propose to use DFSMN as a back-end feature modeling component for speech synthesis.  We investigate the influences of the order and depth in DFSMN to the final system performance. We also try to answer the following two questions: i) \emph{Do we really need to model long-term dependencies in speech synthesis acoustic modeling? }; ii) \emph{Dependency within how long time really matters to synthesized speech quality?}. For comparison, we have trained a strong BLSTM-RNN based speech synthesis system as our baseline system. Experimental results show that DFSMN based system can achieve a comparable voice naturalness to the baseline system while being much smaller in model size and faster in generation speed. Neural networks with the tapped delay structure\cite{tdnn-1,tdnn-2} is another popular feed-forward architecture which can efficiently model the long temporal contexts. The difference between these two models are discussed in \cite{fsmnj}.

The rest of the paper is organized as follows. Details of the proposed DFSMN based speech synthesis system, including the framework, an overview of the compact feed-forward sequential memory networks (cFSMN), and the Deep-FSMN structure is introduced in section 2. Objective experiments and subjective MOS evaluation results are described in Section 3. Conclusion and discussion are given in section 4.

\section{DFSMN based speech synthesis system}
\subsection{System Framework}

The BLSTM based statistical parametric speech synthesis system described in \cite{blstm} is used here as a baseline system. Similar to modern statistical parametric speech synthesis systems, our DFSMN based statistical parametric speech synthesis system is also composed of 3 major parts: the \textit{Vocoder}, the \textit{Front-end}, and the \textit{Back-end}. WORLD\cite{world} is employed in this work as the \textit{Vocoder} to analysis raw speech waveform into spectrum, log F0, and band-periodicity features during training stage. And it is also used to synthesize generated acoustic parameters back to waveform at the synthesis stage. The \textit{Front-end} processes the input text with text normalization and lexical analysis, then feeds the encoded linguistic features as input to neural network training. The \textit{Back-end} in the neural networks based framework is quite straightforward. It establishes mapping between the input encoded linguistic features and the extracted acoustic features with neural networks. During neural network training, different acoustic features streams are learned simultaneously with multi-task training. In this work, DFSMN models with different order and depth are experimented and compared as the \textit{Back-end}, making a direct comparison with the BLSTM based baseline.

\subsection{Compact Feed-forward Sequential Memory Networks}
Compact feed-forward sequential memory networks (cFSMN) is proposed in \cite{cfsmn} as an improvement over standard feed-forward sequential memory networks (FSMN) \cite{fsmn-1,fsmn-2} by combining low-rank matrix factorization. As a variant, cFSMN simplifies FSMN, reduces the number of parameters and speeds up the learning procedure.

\begin{figure}[h]
    \centering
    \includegraphics[width=0.6\columnwidth]{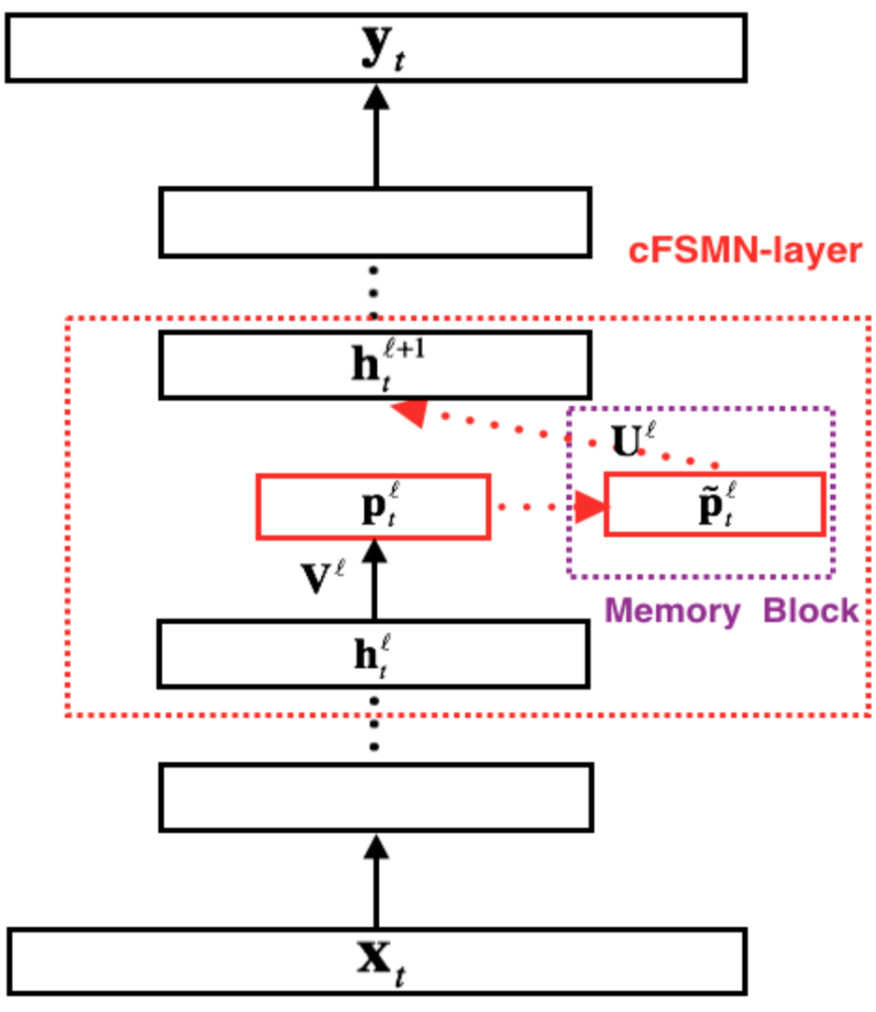}
    \caption{Illustration of cFSMN}
    \label{fig:cfsmn}
\end{figure}

Figure~\ref{fig:cfsmn} gives an illustration of compact feed-forward sequential memory networks (cFSMN). For each cFSMN layer, a linear projection is first applied, the result is then used by the memory block to form an element-wise weighted sum of the history (and the future, if the network is bidirectional) of the layer and finally, the sum is followed by an affine transform and a non-linearity to yield the layer output.

In Equation~\ref{eq:cfsmn-1}, $\mathbf{p}_t^l$ denotes the linear projection of the \textit{l}-th layer. The unidirectional and bidirectional memory blocks are formulated as Equation~\ref{eq:cfsmn-2} and Equation~\ref{eq:cfsmn-3}, respectively. Eventually, the output of the cFSMN layer can be calculated using Equation~\ref{eq:cfsmn-4}. In the equations, $\odot$ denotes element-wise vector multiplication.

\begin{equation}
    \mathbf{p}_t^l = \mathbf{V}^l\mathbf{h}_t^l + \mathbf{b}^l
    \label{eq:cfsmn-1}
\end{equation}

\begin{equation}
    \mathbf{\tilde{p}}_t^l = \mathbf{p}_t^l + \sum_{i=0}^N \mathbf{a}_i^l \odot \mathbf{p}_{t-i}^l
    \label{eq:cfsmn-2}
\end{equation}

\begin{equation}
    \mathbf{\tilde{p}}_t^l = \mathbf{p}_t^l + \sum_{i=0}^{N_1} \mathbf{a}_i^l \odot \mathbf{p}_{t-i}^l + \sum_{j=1}^{N_2} \mathbf{c}_j^l \odot \mathbf{p}_{t+j}^l
    \label{eq:cfsmn-3}
\end{equation}

\begin{equation}
    \mathbf{h}_t^{l+1} = f\left(\mathbf{U}^l \mathbf{\tilde{p}}_t^l + \mathbf{d}^{l}\right)
    \label{eq:cfsmn-4}
\end{equation}

Like RNNs, cFSMN is able to capture long-term information of sequences by tuning the order of memory blocks. Unlike RNNs, cFSMN can be efficiently trained using back-propagation (BP), which is faster and more invulnerable to gradient vanishing problem than doing back-propagation through time (BPTT) with recurrent networks.

\begin{figure*}[htb]
    \centering
    \includegraphics[width=0.9\textwidth]{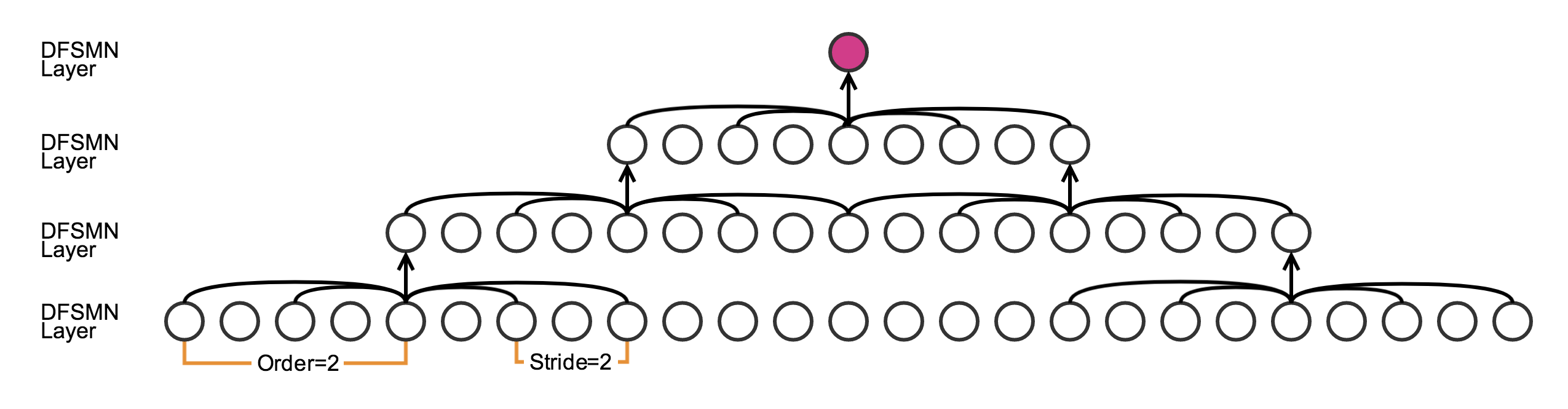}
    \caption{Illustration of effects of order and stride for DFSMN, where in this case $\mathtt{order=2}$ and $\mathtt{stride=2}$}
    \label{fig:order}
\end{figure*}

\begin{table*}[t]
\centering
\caption{Comparison of various DFSMN models in objective measures}
\label{tab:dfsmn}
\resizebox{0.9\textwidth}{!} {
\begin{tabular}{|c|c|c|c|c|c|c|c|c|c|c|}
\hline
\multirow{2}{*}{ID} & \multirow{2}{*}{Network} & \multirow{2}{*}{\#Layers} & \multirow{2}{*}{Order} & \multicolumn{5}{c|}{Objective Measures} & \multirow{2}{*}{\makecell{Size\\(MB)}} & \multirow{2}{*}{\makecell{FLOPS(G)}} \\ \cline{5-9}
\multicolumn{1}{|c|}{} & & & & MCD & F0 RMSE & BAPD & U/V Error & MSE & & \\ \hline
\texttt{BLSTM} & BLSTM & -           & -             & 6.92 & 29.09 & 2.93 & 0.1008 & 0.0273 & 295 & 21.09  \\ \hline\hline
\texttt{A}     & DFSMN & 3+2         & 1,1,1,1       & 7.43 & 33.41 & 3.09 & 0.1074 & 0.0311 & 62  & 4.08  \\ \hline
\texttt{B}     & DFSMN & 3+2         & 2,2,2,2       & 7.33 & 31.96 & 3.03 & 0.1046 & 0.0302 & 62  & 4.08  \\ \hline
\texttt{C}     & DFSMN & 3+2         & 5,5,2,2       & 7.23 & 30.73 & 3.00 & 0.1028 & 0.0294 & 62  & 4.08  \\ \hline
\texttt{D}     & DFSMN & 3+2         & 10,10,2,2     & 7.15 & 30.16 & 2.98 & 0.1019 & 0.0288 & 62  & 4.09  \\ \hline\hline
\texttt{E}     & DFSMN & 6+2         & 10,10,2,2     & 7.11 & 29.91 & 2.97 & 0.1013 & 0.0285 & 87  & 5.35  \\ \hline\hline
\texttt{F}     & DFSMN & 10+2        & 10,10,2,2     & 7.07 & 29.66 & 2.95 & 0.1007 & 0.0282 & 119 & 7.04  \\ \hline
\texttt{G}     & DFSMN & 10+2        & 20,20,2,2     & 6.99 & 29.30 & 2.94 & 0.1004 & 0.0277 & 119 & 7.06  \\ \hline
\texttt{H}     & DFSMN & 10+2        & 40,40,2,2     & 6.92 & 28.92 & 2.91 & 0.0999 & 0.0272 & 120 & 7.10  \\ \hline
\texttt{I}     & DFSMN & 10+2        & 80,80,2,2     & 6.87 & 28.72 & 2.89 & 0.0999 & 0.0269 & 122 & 7.18  \\ \hline
\end{tabular}
}
\end{table*}

\subsection{Deep-FSMN}
Inspired by successful applications of very deep networks in many different fields \cite{vdcnn,resnet,vdcnn-asr-1,vdcnn-asr-2,vdcnn-asr-3,vdcnn-nlp}, one potential improvement over cFSMN is to make the network deeper. We give an overview of Deep-FSMN (DFSMN) \cite{dfsmn} in this section.

\subsubsection{Skip-Connections}
Skip-connections have been proved to be important for exploiting gain from training deep feed-forward neural networks \cite{resnet,highway}. To make cFSMN deeper, it is very interesting to see the performance with the help of skip-connections. DFSMN can not only benefit from the increased representational power of deep network, but may also taking advantage of wider context (history and future) brought by depth indirectly.

For DFSMN, we add skip-connections between consecutive memory blocks $\mathbf{\tilde{p}}_t^{l-1}$ and $\mathbf{\tilde{p}}_t^l$, so that while gradient back-propagating the non-linearities can be bypassed. Equation~\ref{eq:cfsmn-3} now becomes Equation~\ref{eq:dfsmn} where $\mathcal{H}(\mathbf{\tilde{p}}_t^{l-1})$ denotes simple identity mapping of the memory block from the \textit{l-1}-th layer since the memory blocks in DFSMN have the same size.

\begin{equation}
    \mathbf{\tilde{p}}_t^l = \mathcal{H}(\mathbf{\tilde{p}}_t^{l-1}) + \mathbf{p}_t^l + \sum_{i=0}^{N_1} \mathbf{a}_i^l \odot \mathbf{p}_{t-s_1*i}^l + \sum_{j=1}^{N_2} \mathbf{c}_j^l \odot \mathbf{p}_{t+s_2*j}^l
    \label{eq:dfsmn}
\end{equation}

\subsubsection{Order of Memory Blocks}
DFSMN is flexible in terms of context dependency. When sequences are short or latency is essential, the order of memory blocks can be narrowed so that only near history and future are considered by the network. However, when sequences are long or latency is less important, the order of memory blocks can be enlarged to fully utilize long-term dependencies though in the trade of some efficiency.

Apart from order of memory blocks, we also add hyper-parameter \textit{stride} to skip some adjacent frames which in TTS tasks are even more overlapped compared with ASR tasks. On the other hand, by stacking DFSMN layers, the final output is able to see longer history and future hierarchically.

In TTS tasks, long sequences are frequent to see. Consequently, we perform exhaustive search of the order of memory blocks and the depth of DFSMN to try to reach the performance tip of this fantastic neural architecture.

\vspace{-1ex}
\section{Experiments}

\vspace{-1ex}
\subsection{Database}
\vspace{-1ex}
The corpus used in this work is a vocal novel database read by a single Mandarin male speaker. The training set of the corpus contains 38600 utterances (around 83 hours), and another 1400 utterances are left out as validation (around 3 hours). The speech signal is sampled at 16kHz. WORLD is used to extract the 60-dim mel-cepstral features, 3-dim log F0 (static, delta and acceleration), and 11-dim BAP features from the raw recording with frame shift 5ms, and frame length 25ms. Linear interpolation of F0 is done over unvoiced segments before modeling. Unvoiced/voiced identification for every frame is also extracted as one task in the multi-task training, along with the above 3 acoustic streams as separate tasks. The input features for training are the encoded 754 dimensional one-hot and numerical linguistic features obtained from the \textit{Front-end} lexical parsing module given input texts. Both input linguistic features and output acoustic features are normalized to zero-mean and unit-variance before model training. 

\vspace{-1ex}
\subsection{BLSTM baseline}
\vspace{-1ex}
The baseline used in this work is a strong hybrid DNN-BLSTM system. The model is built by stacking one fully-connected layer on the bottom and three BLSTM layers on the top. The fully-connected layer has 2048 hidden units, and each of the BLSTM layers has 2048 cells (1024 for each direction). The BLSTM model is trained using back-propagation through time (BPTT) with stream number 40.

\vspace{-1ex}
\subsection{Training}
\vspace{-1ex}
All models in this work are trained using BMUF \cite{block-ma} optimization on 2 GPUs. The multi-task frame-level MSE (Mean Squared Error) is used as training criteria. The DFSMN models are trained using standard back-propagation (BP) with stochastic gradient descent (SGD), where the batch size is 512. The initial learning rate is 0.0000005, and the learning rate decays by a factor 0.1 when the validation accuracy does not increase enough.

\vspace{-1ex}
\subsection{Objective Evaluations}
\vspace{-1ex}
All DFSMN models are a composition of DFSMN layers and fully-connected layers. One DFSMN layer has 2048 hidden units and 512 projection units, meanwhile each fully-connected layer has 2048 hidden units. In Table~\ref{tab:dfsmn}, column 3 is in form of $N_c+N_d$, which denotes the number of DFSMN layers and fully-connected layers, respectively. Besides, column 4 is in form of $N_1,N_2,s_1,s_2$, indicating that each DFSMN layer looks back $N_1$ frames with stride $s_1$ and looks ahead $N_2$ frames with stride $s_2$. How DFSMN order and stride effects the modeling context window is explained by an example given by Figure~\ref{fig:order}. Here we let DFSMN look back and ahead same frames for simplicity.

\begin{figure}[htb]
    \centering
    \includegraphics[width=\columnwidth]{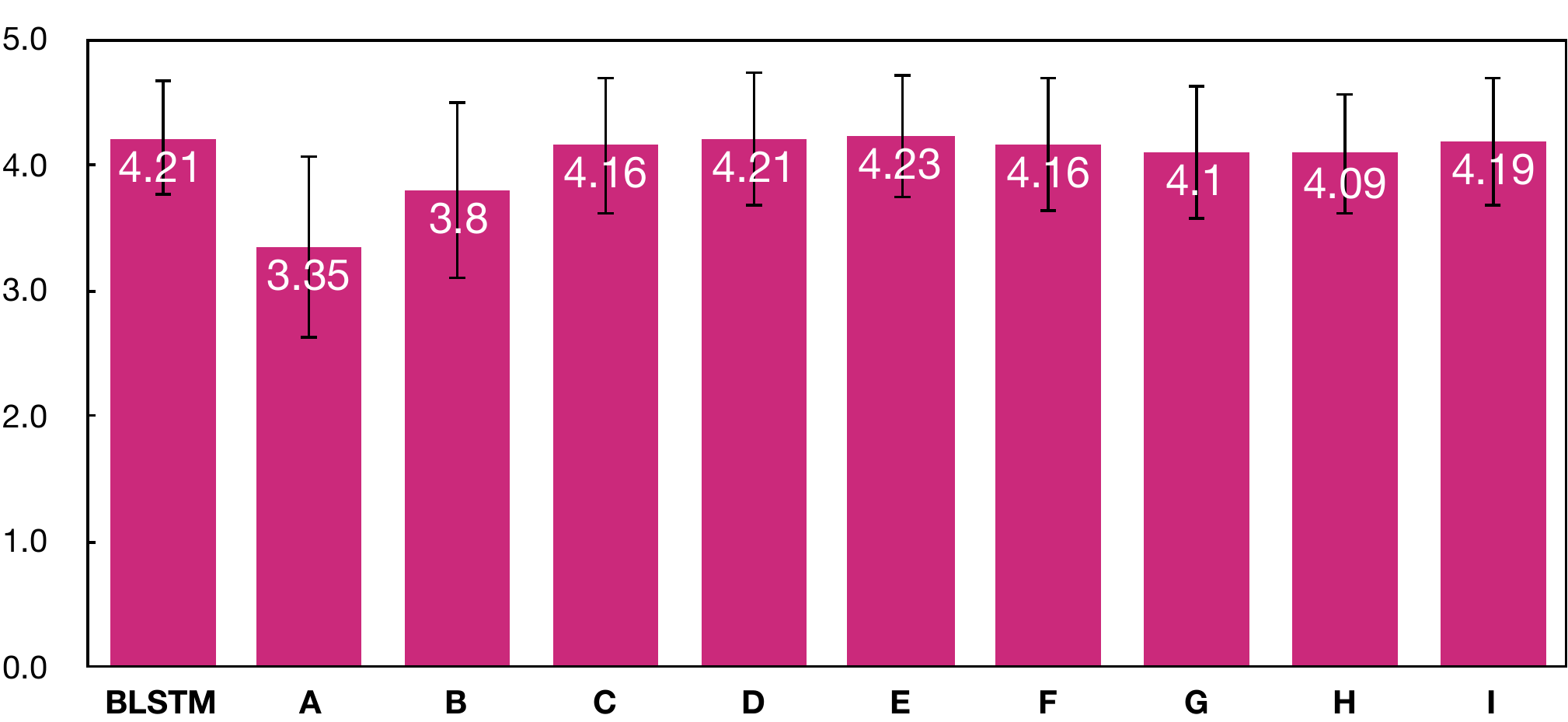}
    \caption{Comparison of various DFSMN models in MOS ($\pm \mathtt{std}$)}
    \label{fig:mos}
\end{figure}

Since the class of FSMN models has not been applied to TTS tasks ever before, we start exploration from a DFSMN model which is shallow and short-sighted. As denoted by \texttt{A} in Table~\ref{tab:dfsmn}, the model contains 3 DFSMN layers and the memory order and stride are both 1, which makes it close to a DNN. In all subsequent experiments, we use a stride of 2, because stride 2 is better than 1 in many informal groups of comparisons, which are not listed in Table~\ref{tab:dfsmn}. We then increase the order of DFSMN models successively ($1 \Rightarrow 2 \Rightarrow 5 \Rightarrow 10$) by fixing the number of DFSMN layers. From \texttt{D} to \texttt{F}, we fix the order and stride as \textit{10,10,2,2} and increase the network depth ($3 \Rightarrow 6 \Rightarrow 10$). Again, from \texttt{F}, we continue to increase the order of DFSMN models ($10 \Rightarrow 20 \Rightarrow 40 \Rightarrow 80$), finally reaching a context window of 3200 frames, approaching max length of utterances in the corpus.

In Table~\ref{tab:dfsmn}, we summarize all the objective measures calculated in this work. These objective measures include root mean squared error (RMSE) of F0 in \textit{Hz}, unvoiced/voiced (U/V) prediction errors, mel-cepstral distortion (MCD) in \textit{dB}, BAP Distortion (BAPD) and the total MSE on normalized acoustic features. The model size and the FLOPS (floating-point operations to generate per second speech) are also recorded. All objective measures are calculated on the 3 hour validation data.

Along with the increment of order and depth, the objective measures consistently drop. With layer depth equals to \textit{10+2}, order and stride equals to \textit{40,40,2,2}, system \texttt{H} beats baseline BLSTM with lower overall MSE, but with less than 1/2 baseline model size, and almost 3 times speech generation speed.

\vspace{-1ex}
\subsection{Subjective MOS tests}
\vspace{-1ex}
Subjective MOS naturalness evaluation is also conducted. Synthesis speech generated from all systems from \texttt{A} to \texttt{I} along with the baseline system are published and evaluated on an internet platform by 40 paid native Mandarin speakers. 20 utterances are generated by each system, and each utterance is listened by 10 different raters in a completely random order. Results are shown in Figure~\ref{fig:mos}. 

The results are very interesting, system \texttt{E} achieves best subjective evaluation performance with MOS score 4.23. Systems get better performance as the order and depth gradually becomes larger until system \texttt{E}, which is as expected. However, after system \texttt{E}, keeping increasing order and depth can not further improve system subjective performance, and the MOS results by these systems seem to oscillate around the best MOS with some noise. Coming back to the two questions raised at the beginning of the paper, the answer is clear. We definitely need to model long-term relations between speech samples in the speech synthesis tasks, and the length of dependency modeling should be around $120 = 6\mathtt{(depth)}\times10\mathtt{(order)}\times2\mathtt{(stride)}$ frames from previous and future samples at maximum, respectively (according to system \texttt{E}). That is a time window of 600ms from each side of the current frame. Longer time window in our experiments seems more like noise and our system shows no further improvement in subjective MOS listening tests.

Besides comparable MOS score with baseline system, system \texttt{E} generates one second speech with 5.35G floating-point operations, which is 4 times faster than BLSTM, making DFSMN very competitive in embedded production environment, where memory saving and computationally efficient is huge advantage.

\vspace{-1ex}
\section{Conclusion and Discussion}
With loops in network structure, RNNs are similar to infinite impulse response (IIR) filter in architecture. On the other hand, FSMN structure is more like finite impulse response (FIR) filter. In theory, One can always use a set of FIR filters to fit a IIR filter. In our paper, we use DFSMN with a range of order and depth to fit the BLSTM structure. With comparable synthesis quality, the proposed DFSMN based TTS system \texttt{E} has much smaller model size, and is computationally much more efficient than the baseline system.

\vspace{-1ex}
\section{Acknowledgements}
The authors would like to thank Zhiying Huang from Alibaba iDST for providing tools for objective distances calculation.


\vfill\pagebreak

\bibliographystyle{IEEEbib}
\ninept
\bibliography{refs}

\end{document}